\RequirePackage{fix-cm}

\documentclass[twocolumn]{svjour3}   

\smartqed  

\usepackage{graphicx,color,multirow,booktabs,bigstrut}

\usepackage[misc]{ifsym}

\usepackage{xpatch,color}

\makeatletter
\def\changeBibColor#1{%
	\in@{#1}{}
	\ifin@\color{red}\else\normalcolor\fi
}

\xpatchcmd\@bibitem
{\item}
{\changeBibColor{#1}\item}
{}{\fail}

\xpatchcmd\@lbibitem
{\item}
{\changeBibColor{#2}\item}
{}{\fail}
\makeatother

\begin{document}

\title{JSRNN: Joint Sampling and Reconstruction Neural Networks for High Quality Image Compressed Sensing}

\author{Chunyan Zeng \textsuperscript{1} \and
		Jiaxiang Ye \textsuperscript{1} \and
        Zhifeng Wang \textsuperscript{2} \and
        Nan Zhao \textsuperscript{1} \and
        Minghu Wu \textsuperscript{1}
}

\institute{\Letter Zhifeng Wang  \at
              \email{zfwang@ccnu.edu.cn}           \\       
   \at
    {$^1$} Hubei Key Laboratory for High-efficiency Utilization of Solar Energy and Operation Control of Energy Storage System, Hubei University of Technology, Wuhan 430068, China
    \at   
    {$^2$} School of Educational Information Technology / Hubei Research Center for Educational Informationization, Central China Normal University, Wuhan 430079, China
}

\date{Received: 11 November 2022}

\maketitle

\begin{abstract}
Most Deep Learning (DL) based Compressed Sensing (DCS) algorithms adopt a single neural network for signal reconstruction, and fail to jointly consider the influences of the sampling operation for reconstruction. In this paper, we propose unified framework, which jointly considers the sampling and reconstruction process for image compressive sensing based on well-designed cascade neural networks. Two sub-networks, which are the sampling sub-network and the reconstruction sub-network, are included in the proposed framework. In the sampling sub-network, an adaptive full connected layer instead of the traditional random matrix is used to mimic the sampling operator. In the reconstruction sub-network, a cascade network combining stacked denoising autoencoder (SDA) and convolutional neural network (CNN) is designed to reconstruct signals. The SDA is used to solve the signal mapping problem and the signals are initially reconstructed. Furthermore, CNN is used to fully recover the structure and texture features of the image to obtain better reconstruction performance. Extensive experiments show that this framework outperforms many other state-of-the-art methods, especially at low sampling rates.
\end{abstract}

\section{Introduction}
Images are ubiquitous in the era of big data, and image processing technology has been developed and applied in many fields \cite{Wang2022ac}. However, in the fields of medical imaging, remote sensing, or others, there are some restrictions when obtaining images, for example, characterizing information by fewer data \cite{Zeng2022b}, massively parallel processing \cite{Wang2021}, abundant redundant information in cloud storage \cite{Zeng2020a}, less bandwidth of information transmission \cite{Wang2017}, and space required for signal storage \cite{Zeng2022}, etc. Fortunately, the emergence of the theory of Compressed Sensing (CS) \cite{Zeng2021c} has successfully solved these restrictions.

CS theory, which obtains sampling information far below the frequency of Nyquist sampling and recovers the original signal from the sampling information with high probability, breaks through the limitation of the traditional sampling law \cite{Wang2022ac,Wang2015a}. CS theory has a profound impact on statistics, information theory, coding theory, etc. People can use less information to represent more data without being affected by the signal bandwidth and storage. However, the traditional compressed sensing reconstruction algorithms are difficult to reconstruct images in real-time. Fortunately, with the wide application of deep learning technology, the deep learning based reconstruction algorithms are several orders of magnitude faster than the traditional reconstruction algorithms. In the field of computer imaging, the University of Glasgow has successfully designed a new single-pixel camera based on compressed sensing \cite{high}. It demonstrates the application of deep learning with convolutional auto-encoder networks to recover real-time 128$\times $128-pixel video at 30 frames-per-second from a single-pixel camera sampling at a compression ratio of 2$\% $. This has taken a substantial step to replace the traditional camera in practice. This paper focuses on the end-to-end framework of sampling and reconstruction for real-time and high-quality signal reconstruction at a low sampling rate. The contributions of this paper are as follows$:$

(1). By cascading SDA and CNN modules, we propose a CS framework with high reconstruction performance especially at a low sampling rate. SDA is used for initial reconstruction, and CNN is applied for further effective and fast reconstruction.

(2). We adopt a nonlinear adaptive sampling network to learn and characterize the signal structure through the data-driven method, and jointly consider sampling and reconstruction, which solves the problem that the traditional Gaussian random matrix has insufficient ability to represent specific data.

(3). Experiments on natural images show that the reconstruction performance of our cascading network based on nonlinear measurements is better than baseline with the same iterations.

The rest of the paper is organized as follows: Section \ref{2} gives an overview of the related work in the area of CS. Section \ref{3} describes the proposed method by a step-by-step detailed explanation. Experiments are presented in Section \ref{4}, while the conclusions are given in Section \ref{5}.

\section{Related work} \label{2}
\subsection{The classical framework of compressed sensing}

The CS theory can be expressed as:
\begin{equation}
y = \Phi x
\end{equation}
Where $y \in {R^M}$ is a measurement vector, $x \in {R^N}$ represents the original signal, and $M \ll N$, $\Phi  \in {R^{M \times N}}$ is a measurement matrix.

CS theory has two important components: sampling and reconstruction. Sampling needs to retain enough information for reconstruction. The process of obtaining $y$ from $x $ can be understood as obtaining the low-dimensional representation signal. And reconstructing the original signal $x $ from $y $ can be understood as solving an inverse problem from a set of undersampled linear measurements. 

The design of the sampling matrix needs to meet the restricted isometry property (RIP) \cite{canh2021restricted}, and the widely used random measurement matrix includes Gaussian, Bernoulli, and so on. However, the random matrix has large memory requirements and high computational complexity which restricts the applications of CS. More importantly, the random matrix is not designed for specific signals, and measurements obtained by it can't provide sufficient reconstruction information. The Toeplitz matrix and polynomial matrix are utilized for sampling recently \cite{liu2021multi}, which reduces the computation cost but leads to worse reconstruction quality than that with a random matrix. Some researchers have designed the sampling matrix for a specific signal to obtain a better reconstruction effect. In \cite{7149270}, Gao et al. utilize the local smooth property to obtain a local structural sampling matrix, but this is only useful for a certain type of signal, and often fails for others. In this paper, we use a neural network to sample the data and adaptively obtain the implicit information needed for reconstruction, which reduces the computational complexity, storage space, and also takes a better effect on reconstruction.

The traditional reconstruction methods can be divided into three categories: convex relaxation algorithms, greedy algorithms, and Bayesian algorithms. The convex relaxation algorithms \cite{saha,Li2020} approximate the non-convex  ${l_0}$ norm with a versatile mixed norm. The approximated problem can be solved by the standard optimization method. The typical algorithms using convex relaxation are Basis Pursuit (BP) \cite{zhang2021matrix} algorithm, Gradient Projection for Sparse Reconstruction (GPSR) \cite{liu2018} algorithm and Interior Point Method (IPM) \cite{lin2020admm} algorithm. Greedy algorithms \cite{Zhang2020,Tirer2020} update the estimated signal support set iteratively to approximate the target signal, which includes two basic steps: atomic selection and signal updating estimation. Typical greedy algorithms mainly include Matching Pursuit (MP)\cite{mp} algorithm, Orthogonal Matching Pursuit (OMP) \cite{zarei2021automatic} algorithm and Compressive Sampling Matching Pursuit (CoSaMP)\cite{Tirer2020} algorithm. Based on statistics, Bayesian algorithms \cite{Zhao2020,xu2021interpolation,Montoya-Noguera2019} adopt the prior probability density distribution function of the signal to obtain the maximum posterior probability and estimate the reconstruction error range, and further reconstruct the original signal. Besides, D-AMP \cite{Metzler2016} algorithm is based on denoising and probability map and predicts the next iteration through state evolution. NLR-CS \cite{Dong2014} algorithm uses the nonlocal similarity of the image itself to construct a low-rank matrix model and realize image reconstruction.

The advantage of traditional reconstruction methods is that they are based on interpretable prior knowledge, such as the sparsity of signal structure. Furthermore, when the reconstruction problem based on the ${l_0}$ norm minimum is modeled as a convex optimization problem, there is a theoretical convergence guarantee. However, the fastest of these algorithms using precise prior knowledge of signals is too slow to apply to real-world signals, and these algorithms do not utilize any training data as a resource for feature extraction. Besides, the design of the measurement matrix is very important for reconstruction performance, but traditional CS reconstruction and sampling are designed separately. Sampling matrices are usually random matrices and do not be designed jointly with reconstruction algorithms, thus improving the reconstruction performance.

\subsection{Deep learning based compressed sensing}
The rapid development of deep learning provides researchers with a new solution for CS reconstruction. The end-to-end reconstruction greatly reduces the computational complexity of CS reconstruction, training by providing enough data to the reconstruction neural network, the reconstruction performance is comparable to traditional algorithms. We can divide the algorithms of deep learning based compressed sensing (DCS)  reconstruction into two categories. The first category is data-driven algorithms based on prior knowledge. And these algorithms can be divided into two sub-categories. The first subcategory treats the neural network as a black box that performs some function \cite{tramel2018deterministic}, such as computing a posterior probability. The second subcategory explicitly expands the iterative algorithm into a neural network and then fine-tunes it with training data. For example, ISTA-Net \cite{Zhang2018}, AMP-Net \cite{zhang2020amp} and ADMM-CSNet \cite{ADMM-CSNet} algorithms unfold the traditional ISTA, AMP and ADMM algorithms respectively. The second category is a data-driven algorithm without using any signal prior knowledge. Mousavi et al. \cite{SDA} design SDA to reconstruct the image, which is the first image reconstruction algorithm using the deep learning method. The DeepInverse \cite{Mousavi2017}, DR2-Net \cite{Yao_2019}, ReconNet \cite{recon} and CS-Net \cite{Shi2020} adopt convolutional layers or residual blocks to build a deep learning framework, and the reconstruction effects of these algorithms are better compared with  \cite{SDA}. Compared with traditional reconstruction algorithms, the deep learning-based reconstruction algorithms not only have the comparable accuracy, but also run thousands of times faster. If the above problems can be solved, CS can be widely used in many tasks, such as image processing \cite{Wang2021,Wang2015a}, speaker recognition \cite{Wang2021m,Wang2020h,Zeng2018}, source identification \cite{Zeng2021a,Zeng2020,Zeng2021b,Wang2015b}, audio forensics \cite{Zeng2022a,Wang2022t,Wang2018a}, information understanding \cite{Lyu2022}, and other fields.

\subsection{Motivations}
The research of CS has been constrained by two significant challenges. One is that traditional sampling and reconstruction methods rely on the prior assumption of signal sparsity, however the actual data is not completely sparse in the transform domain. Therefore, how to learn the complex signal structure of real signals to improve the reconstruction performance remains to be solved. Another is that at low measurement rates, the reconstructed images by traditional methods are of poor quality and can't even be used for image understanding tasks.

To address the first challenge, Baraniuk et al. in \cite{SDA} use SDA directly to sample and reconstruct signals. This framework is inspired by SDA, which applies the encoder network to deal with the problem of data dimensionality reduction and the decoder network to recover the data \cite{SDA}. Nevertheless, the image quality reconstructed by SDA is not significantly improved compared with the traditional algorithms. To address the second challenge, Reconnet \cite{recon} combines a fully connected layer without activation function and six convolutional layers to reconstruct images at a low sampling rate. However, this preliminary recovering of the measurement signal is not well-designed, which uses only one layer of full connection layer to complete the increase of signal dimension.

In this paper, we regard the sampling and reconstruction of CS as an end-to-end process of signal coding and decoding by data-driven without sparsity as a priori knowledge. So we use the first layer of SDA, i.e. the full connection layer, to adaptively sample and learn the structure features, and then use the other layers of SDA to initially reconstruct the image. Because CNN is good at learning edges and details of image signals, we cascade CNN after SDA and further enhance the quality of image reconstruction. Through the initial reconstruction of SDA and further reconstruction of CNN, the cascaded network has much better reconstruction performance than Reconnet with the same network depth.

\begin{figure*}[!htp]
	\centering
	{\includegraphics[width=6.7in]{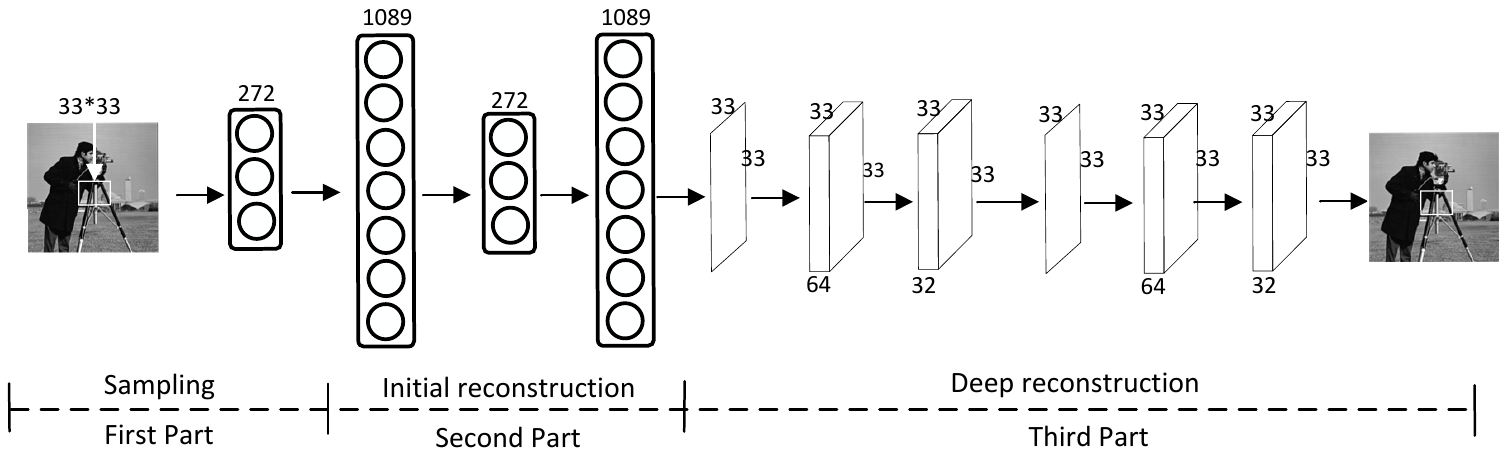}}
	\caption{{Overview of the proposed framework.}
		This figure shows a schematic diagram of the network structure with a measurement rate of 0.25, the measurement rate is changed by the number of neurons in the first layer.}
\end{figure*}

\section{Proposed method} \label{3}
In order to adaptively sense signals and get a better reconstruction performance at a low sampling rate, this paper proposes a new signal acquisition and reconstruction framework. The framework we proposed contains the sampling network, initial reconstruction, and deep reconstruction network. The overall block diagram of the sampling and reconstruction algorithm is shown in Fig. 1. The first part shows the sampling network, which consists of one fully connected layer. The second part is the initial reconstruction network, consisting of a 3-layer SDA network, which boosts the dimension of the signal and initially reconstructs the signal. The third part is the deep reconstruction network, which consists of a 6-layer convolutional neural network without any pooling layer.

\subsection{The Sampling Network}
The sampling network is the first layer of the network proposed in this paper, in which the neurons are fully connected with the original signal. The key information that is needed in the reconstruction work is adaptively acquired by the sampling network, which is trained jointly with the reconstruction networks. The neurons number of the fully connected layer varies with the measurement rate.

In the sampling network, the original signal (the $33*33$ image block in Fig. 1) is rearranged into a column to form the high-dimensional original signal $x \in {R^M}$. After sampling, the measurement signal $y \in {R^M}$ is obtained. The process can be expressed as:
\begin{equation}
y = {\rm T}\left( {{W_1}x + {b_1}} \right)
\end{equation}
Where ${W_1}$ and ${b_1}$ represent the weight and bias of the sampling layer,  ${\rm T}$ represents the activation function ReLU, $x \in {R^N}$ represents the input original signal, and $y \in {R^N}$ represents the measurement signal.

\subsection{Initial reconstruction}
In the second part of the framework, we need to find a mapping from the low to the high dimension for boosting pixels and initially reconstructing. We leverage a 3-layer SDA neural network to achieve this goal. SDA has an excellent performance in dealing with signal dimensional problems. And at the same time, the SDA can initially reconstruct the original image while improving the signal dimension.

As shown in Fig. 1, in the 2-4 layers, the reconstruction network takes the measurement signal as input and outputs the initial reconstruction signal. Each layer of the network can be expressed as:
\begin{equation}
{x_i} = {\rm T}\left( {{W_i}{x_{i - 1}} + {b_i}} \right)
\end{equation}
Where ${x_i}$ represents the output of the $i$-th layer and $i = 2,3,4$, ${x_1}$ especially represents the measurement signal and ${x_1} = y$. ${W_i}$ is the weight matrix of the $i$-th layer, ${b_i}$ is the bias of the $i$-th layer,  ${\rm T}$ represents the nonlinear activation function ReLU. The dimension of ${x_i}$ is related to the number of neurons in the layer of SDA. We set the numbers of neurons in the three-layer SDA to 1089, 272, and 1089 respectively, due to the block size of the original image 33*33. ${x_4}$ is then reshaped into an image block as the preliminary reconstructed image.

\subsection{Deep reconstruction}
Since the preliminary reconstruction image blocks are still far different from the original image blocks, we need to further narrow the gap between them. We design a convolutional neural network composed of 6 convolutional layers to achieve the goal. In order not to change the size of the feature map, we give up the downsampling pooling operation and appropriately use the zero-padding operation.

Inspired by Reconnet \cite{recon}, there are six layers in the deep reconstruction network, and the parameters of the first three layers are the same as the last three layers. The size settings of the six convolutional layers are 11*11, 1*1, 7*7, 11*11, 1*1, 7*7 to keep the size of the output feature map remaining 33*33, accordingly the number of feature maps output at each layer are 64, 32, 1, 64, 32, 1. Except for the last layer, the other five layers apply the nonlinear activation function ReLU. The last layer is the output layer of the network, and the output is the reconstructed image blocks.

\subsection{The loss function and training process}
In order to reduce the parameters and balance the computational complexity of the network and reconstruction performance, we divide the original image into 33*33 sub-image blocks. The numbers of image blocks divided by images in the dataset are not equal, and it depends on the size of each image. The total number of image blocks in the database is $S$.

Through the joint training of the sampling layer and the reconstruction network, the reconstruction network can guide the optimization direction of the sampling layer and enable the sampling layer to obtain the information needed for reconstruction adaptively.

The Adam and SGD optimization algorithms are applied to optimize all parameters ${W_L} = \left\{ {{W_i},{b_i}} \right\}$ (we both use two optimization algorithms to optimize the network separately and select the network with lower loss as the final sampling reconstruction network), ${W_i}$ and ${b_i}$ respectively represent the weight and bias of each layer. The loss function is as follows:
\begin{equation}
L\left( {{W_L}} \right) = \frac{1}{S}\sum\limits_i^S {\left\| {{\rm T}\left( {{x_i},{W_L}} \right) - {x_i}} \right\|} _2^2\
\end{equation}
Where ${\rm T}\left( {{x_i},{W_L}} \right) $ represents the output of the network.

\section{Experimental results and discussion} \label{4}
\subsection{Training and testing dataset}

\begin{table}[htbp]
	\centering
	\caption{The mean PSNR at different rates with pre-train and without pre-train.}\label{tab1}
	\begin{tabular}{ccccc}
		\hline
		\multirow{2}[4]{*}{Methods} & \multicolumn{4}{c}{Measurement Rates} \bigstrut\\
		\cline{2-5}          & 0.25  & 0.1   & 0.04  & 0.01 \bigstrut\\
		\hline
		With pre-train & \textbf{26.43} & \textbf{23.20}  &\textbf{ 22.00}    & 13.07 \bigstrut[t]\\
		Without pre-trian & 25.71 & 20.34 & 21.34 & \textbf{17.34} \bigstrut[b]\\
		\hline
	\end{tabular}%
\end{table}%

\begin{table}[htbp]
	\centering
	\caption{The parameters of the proposed framework.}\label{tab2}
	\begin{tabular}{ccccc}
		\hline
		\multirow{2}[4]{*}{Parameters} & \multicolumn{4}{c}{Measurement Rates} \bigstrut\\
		\cline{2-5}          & 0.25  & 0.1   & 0.04  & 0.01 \bigstrut\\
		\hline
		SL    & 272   & 108   & 43    & 10 \bigstrut[t]\\
		FC1   & 1089  & 1089  & 1089  & 1089 \\
		FC2   & 272   & 272   & 272   & 272 \\
		FC3   & 1089  & 1089  & 1089  & 1089 \\
		CONV1 & 11*11*64 & 11*11*64 & 11*11*64 & 11*11*64 \\
		CONV2 & 1*1*32 & 1*1*32 & 1*1*32 & 1*1*32 \\
		CONV3 & 7*7*1 & 7*7*1 & 7*7*1 & 7*7*1 \\
		CONV4 & 11*11*64 & 11*11*64 & 11*11*64 & 11*11*64 \\
		CONV5 & 1*1*32 & 1*1*32 & 1*1*32 & 1*1*32 \\
		CONV6 & 7*7*1 & 7*7*1 & 7*7*1 & 7*7*1 \bigstrut[b]\\
		\hline
	\end{tabular}%
\end{table}%

\begin{table*}[htbp]
	\centering
	\caption{PSNR values in dB for the test images by different algorithms at different measurement rates.} \label{tab3}
	\begin{tabular}{cccccc|cccccc}
		\hline
		\multirow{2}[4]{*}{Data} & \multirow{2}[4]{*}{Methods} & \multicolumn{4}{c|}{Measurement Rates} & \multirow{2}[4]{*}{Data} & \multirow{2}[4]{*}{Methods} & \multicolumn{4}{c}{Measurement Rates} \bigstrut\\
		\cline{3-6}\cline{9-12}          &       & 0.25  & 0.10  & 0.04  & 0.01  &       &       & 0.25  & 0.10  & 0.04  & 0.01 \bigstrut\\
		\hline
		\multirow{7}[2]{*}{Monarch} & NLR-CS & 25.91  & 14.59  & 11.62  & 6.38  & \multirow{7}[2]{*}{Finstones} & NLR-CS & 22.43  & 12.18  & 8.96  & 4.45  \bigstrut[t]\\
		& D-AMP & 26.39  & 19.00  & 14.57  & 6.20  &       & D-AMP & 25.02  & 16.94  & 12.93  & 4.33  \\
		& SDA   & 23.54  & 20.95  & 18.09  & 15.31  &       & SDA   & 20.88  & 18.40  & 16.19  & 13.90  \\
		& Reconnet & 22.84  & 20.55  & 17.77  & 15.01  &       & Reconnet & 20.73  & 18.45  & 15.73  & 13.79  \\
		& DR2-Net & 27.95  & 23.10  & 18.93  & 15.33  &       & DR2-Net & 26.19  & 21.09  & 16.93  & 14.01  \\
		& ISTA-Net & \textbf{32.54 } & \textbf{25.58 } & 19.40  & 14.99  &       & ISTA-Net & \textbf{29.37 } & \textbf{23.39 } & 17.43  & 14.00  \\
		& Ours  & 26.02  & 23.76  & \textbf{21.67 } & \textbf{17.25 } &       & Ours  & 24.00  & 21.44  & \textbf{18.89 } & \textbf{15.98 } \bigstrut[b]\\
		\hline
		\multirow{7}[2]{*}{Parrots} & NLR-CS & 26.53  & 14.14  & 10.59  & 5.11  & \multirow{7}[2]{*}{Foerman} & NLR-CS & 35.73  & 13.54  & 9.06  & 3.91  \bigstrut[t]\\
		& D-AMP & 26.86  & 21.64  & 15.78  & 5.09  &       & D-AMP & 35.45  & 13.54  & 9.06  & 3.91  \\
		& SDA   & 24.48  & 22.13  & 20.37  & 17.70  &       & SDA   & 28.39  & 26.43  & 23.62  & 20.07  \\
		& Reconnet & 24.27  & 22.34  & 19.93  & 17.14  &       & Reconnet & 28.60  & 26.81  & 23.63  & 19.61  \\
		& DR2-Net & 28.73  & 23.94  & 21.16  & 18.01  &       & DR2-Net & 33.53  & 29.20  & 25.34  & 20.59  \\
		& ISTA-Net & \textbf{31.42 } & \textbf{26.21 } & 22.24  & 17.90  &       & ISTA-Net & \textbf{38.23 } & \textbf{32.78 } & 25.76  & 20.21  \\
		& Ours  & 27.33  & 25.21  & \textbf{23.09 } & \textbf{20.50 } &       & Ours  & 31.35  & 29.44  & \textbf{27.48 } & \textbf{23.34 } \bigstrut[b]\\
		\hline
		\multirow{7}[2]{*}{Cameraman} & NLR-CS & 24.88  & 14.18  & 11.04  & 5.98  & \multirow{7}[2]{*}{Lena} & NLR-CS & 29.39  & 15.30  & 11.61  & 5.95  \bigstrut[t]\\
		& D-AMP & 24.41  & 20.35  & 15.11  & 5.64  &       & D-AMP & 28.00  & 22.51  & 16.52  & 5.73  \\
		& SDA   & 22.77  & 22.15  & 19.32  & 17.11  &       & SDA   & 25.89  & 23.81  & 21.18  & 17.84  \\
		& Reconnet & 22.25  & 20.79  & 18.75  & 16.88  &       & Reconnet & 25.44  & 23.53  & 21.09  & 17.54  \\
		& DR2-Net & 25.62  & 22.46  & 19.84  & 17.08  &       & DR2-Net & 29.42  & 25.39  & 22.13  & 17.97  \\
		& ISTA-Net & \textbf{28.61 } & \textbf{23.46 } & 20.27  & 17.26  &       & ISTA-Net & \textbf{32.30 } & \textbf{27.44 } & 22.40  & 18.29  \\
		& Ours  & 24.93  & 23.31  & \textbf{21.64 } & \textbf{19.17 } &       & Ours  & 28.37  & 26.22  & \textbf{24.41 } & \textbf{20.53 } \bigstrut[b]\\
		\hline
		\multirow{7}[2]{*}{Fingerprint} & NLR-CS & 23.52  & 12.81  & 9.66  & 4.85  & \multicolumn{1}{c}{\multirow{7}[2]{*}{Mean}} & NLR-CS & 26.91  & 13.82  & 10.36  & 5.23  \bigstrut[t]\\
		& D-AMP & 25.17  & 17.15  & 13.82  & 4.66  &       & D-AMP & 27.32  & 20.44  & 15.00  & 5.07  \\
		& SDA   & 24.28  & 20.29  & 16.87  & 14.83  &       & SDA   & 24.31  & 21.88  & 19.37  & 16.68  \\
		& Reconnet & 23.88  & 20.26  & 16.60  & 14.71  &       & Reconnet & 24.00  & 21.81  & 19.07  & 16.38  \\
		& DR2-Net & 27.65  & 22.03  & 17.40  & 14.73  &       & DR2-Net & 28.44  & 23.89  & 20.25  & 16.82  \\
		& ISTA-Net & \textbf{28.10 } & 22.45  & 17.31  & 14.78  &       & ISTA-Net & \textbf{31.51 } & \textbf{25.90 } & 20.69  & 16.78  \\
		& Ours  & 27.10  & \textbf{24.99 } & \textbf{19.46 } & \textbf{16.06 } &       & Ours  & 27.01  & 24.91  & \textbf{22.37 } & \textbf{18.97 } \bigstrut[b]\\
		\hline
	\end{tabular}%
\end{table*}%

\begin{table*}[htbp]
	\centering
	\caption{Structural Comparison.}
	\begin{tabular}{ccccccccc}
		\hline
		\multirow{2}[4]{*}{Data} & \multicolumn{2}{c}{MR=0.01} & \multicolumn{2}{c}{MR=0.04} & \multicolumn{2}{c}{MR=0.10} & \multicolumn{2}{c}{MR=0.25} \bigstrut\\
		\cline{2-9}          & Ours  & CNNs+SDA & Ours  & CNNs+SDA & Ours  & CNNs+SDA & Ours  & CNNs+SDA \bigstrut\\
		\hline
		Monarch & \textbf{17.25 } & 16.55  & \textbf{21.67 } & 20.77  & \textbf{23.76 } & 23.07  & \textbf{26.02 } & 25.49  \bigstrut[t]\\
		Parrots & \textbf{20.50 } & 19.89  & \textbf{23.09 } & 22.85  & \textbf{25.21 } & 24.60  & \textbf{27.33 } & 27.18  \\
		Barbara & \textbf{20.31 } & 20.23  & \textbf{22.85 } & 22.84  & \textbf{23.70 } & 23.57  & \textbf{24.40 } & 24.39  \\
		Boats & \textbf{20.44 } & 20.15  & \textbf{24.32 } & 24.27  & \textbf{26.84 } & 26.38  & \textbf{29.03 } & 28.82  \\
		Cameraman & \textbf{19.17 } & 18.87  & \textbf{21.64 } & 21.58  & \textbf{23.31 } & 23.02  & \textbf{24.93 } & 24.71  \\
		Fingerprint & \textbf{16.06 } & 15.89  & \textbf{19.46 } & 19.87  & \textbf{24.99 } & 24.11  & 27.10  & \textbf{27.45 } \\
		Flinstones & \textbf{15.98 } & 15.53  & \textbf{18.89 } & 18.64  & \textbf{21.44 } & 20.95  & \textbf{24.00 } & 23.96  \\
		Foreman & \textbf{23.34 } & 22.60  & \textbf{27.48 } & 26.98  & \textbf{29.44 } & 28.82  & \textbf{31.35 } & 31.45  \\
		House & \textbf{21.71 } & 21.03  & \textbf{25.94 } & 25.73  & \textbf{28.54 } & 28.02  & \textbf{30.35 } & 30.35  \\
		Lena256 & \textbf{20.53 } & 19.99  & \textbf{24.41 } & 24.12  & \textbf{26.22 } & 25.58  & \textbf{28.37 } & 28.07  \\
		Peppers256 & \textbf{18.80 } & 18.46  & \textbf{22.43 } & 22.41  & \textbf{24.45 } & 24.18  & \textbf{26.55 } & 26.39  \bigstrut[b]\\
		\hline
	\end{tabular}%
	\label{tab4}%
\end{table*}%

In the preparation of the training data, we get the same 91 natural images as in Recconnet\cite{recon} and convert all the images into grayscale images. And we get 11 grayscale images, such as Lena, Monach, etc, as test data. In order to reduce the storage capacity and the computational complexity of the network, we need to take the image blocking strategy. The training images are divided into several $33*33$ sub-image blocks in an overlapping mode by the stride step size of 12-pixel. By contrast, the test images are also be blocked but in non-overlapping mode. Because we don't need to reconstruct a part of an image for more than one time, and this can reduce the reconstruction time of a whole image. It should be noted that when the last part of the image in the horizontal or vertical direction is less than 33*33 for blocking, the zero-padding operation needs to be performed at the edge of the image.

\subsection{Training strategies}
It is obvious that the parameters in the convolutional layers are fewer than that of fully connected layers. If the fully connected layers and convolution layers are trained together, the gradient will disappear when the weights of the fully connected layers are adjusted in the backpropagation after a certain number of iterations. So we pre-train the full connection layers, and give the parameters of the full connection layers as the starting value to the cascade network, and then fine-tune the whole network.

According to the conventional training way, we may complete the training of the 10-layer network together, and perform end-to-end training by setting the parameter parameters of each layer to zero or giving some starting value by Gaussian random initialization. However, our network structure is formed by cascading fully connected layers and convolutional layers. In order to prevent the phenomenon mentioned above, we cascade the first four layers of the framework to take a pre-train.

We compare PSNR results of test images with pre-trained and non-pre-trained methods at different measurement rates. The average PSNR value of the test results in Table 1 can be seen: in the case of a high measurement rate, the pre-training can significantly improve the reconstruction performance of the framework, however, when the measurement rate is low, the pre-training reduce the reconstruction quality. That is because as the measurement rate is lower, the number of neurons in the sampling network is extremely little. For example, at MR=0.01, the sampling layer is composed of 10 neurons, and excessive training lead to over-fitting of the sampling layer. In summary, we compare PSNR results of test images with pre-trained and non-pre-trained methods at different measurement rates. In these two modes, we choose a network structure with a lower loss as the final sampling reconstruction network.

\subsection{Implementation details}
We complete the training of the proposed framework in Caffe. When training the proposed framework, we set the number of iterations to 1000000 and the learning rate to 0.001. We train on the server host with i7-CPU and GT-1080ti-GPU. It takes about 15 hours to complete a framework training at one measurement rate. The structural details of the proposed frameworks are shown in Table 2.

\subsection{PSNR comparison results}
To compare the proposed frameworks with state-of-the art algorithms, we choose two traditional iterative algorithms NLR-CS \cite{Dong2014}, D-AMP \cite{Metzler2016}, and four latest deep learning-based algorithms SDA \cite{SDA}, Reconnet \cite{recon}, DR2-Net \cite{Yao_2019}, and ISTA-Net \cite{Zhang2018}.

From Table 3 we can see that: in the case of low measurement rates(0.01 and 0.04), our frameworks have achieved the best reconstruction effect than others, and the PSNR values are much higher than that of the traditional iterative algorithms. Further, in Fig. 2, we show the histogram of mean PSNR results at different measurement rates, and our frameworks have achieved an excellent reconstruction effect more intuitively. At the high measurement rates(0.10 and 0.25), our frameworks is obviously superior to SDA, Reconnet, and traditional iterative algorithms, but is inferior to ISTA-Net. The reason is that our method only uses 6-layer convolutions in the deep reconstruction phase, while ISTA-Net uses a 54-layer network, which is far deeper than our network.

\begin{figure}[h!]
	\caption{{Performance comparison of reconstruction methods based on deep learning.}}
	\includegraphics[width=3.2in]{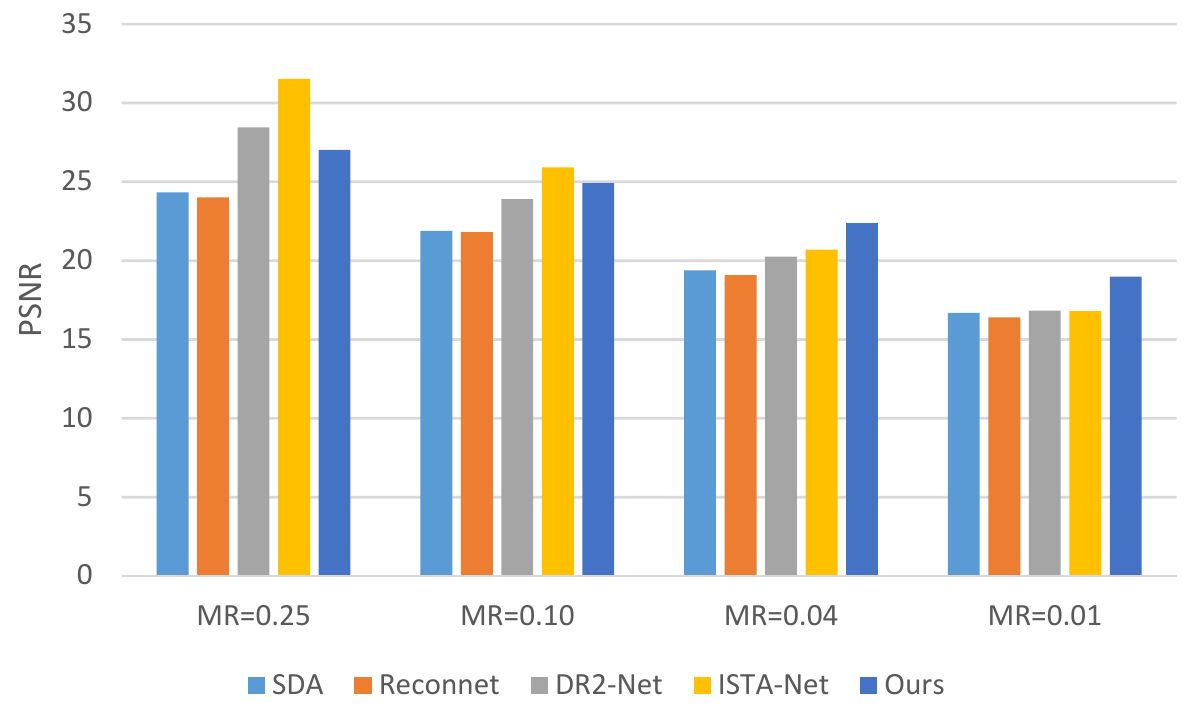}
	\label{fig2}
\end{figure}

\subsection{Structural comparison}
In order to find a more effective cascade structure and compare it fairly with Reconnet, we cascade 3 layers of fully connected layer in the rear of Reconnet network, called RecSDA, whose network depth is the same as the reconstruction part of the proposed framework, and all of which is 9 layers. The numbers of parameters of the two networks are the same, and the test results are shown in Table 4. The proposed framework is much better than the RecSDA in all measurement rates, which shows our structure is effective and better.

\begin{table}[htbp]
	\centering
	\caption{Time complexity.}
	\begin{tabular}{ccccc}
		\hline
		\multirow{2}[4]{*}{Methods} & \multicolumn{4}{c}{Measurement Rates} \bigstrut\\
		\cline{2-5}          & 0.25  & 0.1   & 0.04  & 0.01 \bigstrut\\
		\hline
		NLR-CS [19] & 314.852 & 305.703 & 300.666 & 314.176 \bigstrut[t]\\
		D-AMP [18] & 27.764 & 31.849 & 34.207 & 54.643 \\
		SDA [24]  & \textbf{0.0034} & \textbf{0.0034} & \textbf{0.0033} & \textbf{0.0031} \\
		Reconnet [27] & 0.0104 & 0.01  & 0.0099 & 0.0103 \\
		DR2-Net [26] & 0.0326 & 0.0314 & 0.0317 & 0.0317 \\
		ISTA-Net [21] & 0.0188 & 0.0197 & 0.0182 & 0.0207 \\
		Ours  & 0.0112 & 0.0108 & 0.0097 & 0.0101 \bigstrut[b]\\
		\hline
	\end{tabular}%
	\label{tab5}%
\end{table}%

\subsection{Time complexity}
We test the running time of different algorithms, and it is shown in Table 5. It is obvious that the deep learning-based reconstruction algorithms are an order of magnitude faster than the traditional iterative algorithms, and the SDA algorithm has the fastest speed. That is because as the number of layers in the network increase, the time required for forwarding propagation will also increase. Fig. 3 compares different algorithms in terms of PSNR and time consumption, and it shows that five methods based on deep learning have achieved real-time reconstruction of images.

\begin{figure}[h!]
	\caption{{Time complexity comparison.}
		This figure shows the time consumption comparison between the traditional reconstruction algorithm and the deep learning-based reconstruction algorithm with a measurement rate of 0.04. The deep learning algorithm can achieve real-time reconstruction, but our algorithm can obviously achieve better performance.}
	\includegraphics[width=3.2in]{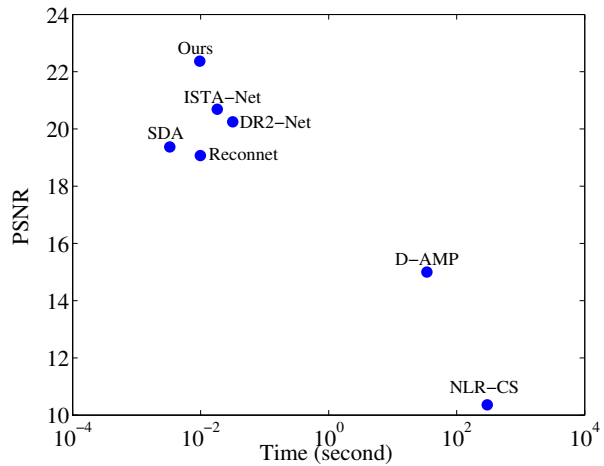}
	\label{fig3}
\end{figure}

\section{Conclusion} \label{5}
In this paper, we have designed a new framework for joint sampling and reconstruction of images by cascading SDA and CNN, which can reconstruct signals accurately at low sampling rates. The SDA boosts the dimensionality and initially reconstruct the signals and the CNN further improve the reconstruction quality. Experiments suggest that the proposed framework is faster and more efficient compared with the traditional iterative algorithms and get better reconstruction results at low sampling rates than the other deep learning-based methods in a similar structure. 

\section*{Acknowledgements}
This work was supported by National Natural Science Foundation of China (No.61901165, 61501199), Science and Technology Research Project of Hubei Education Department (No. Q20191406), Hubei Natural Science Foundation (No. 2017CFB683), Hubei Research Center for Educational Informationization Open Funding (No. HRCEI2020F0102), and Self-determined Research Funds of CCNU from the Colleges’ Basic Research and Operation of MOE (No. CCNU20ZT010).

\bibliographystyle{spmpsci}
\bibliography{bmc_article,MyWork}

\end{document}